\documentclass[11pt]{article}
\usepackage[margin=1in]{geometry}
\usepackage{amsmath,amssymb}
\usepackage{graphicx}
\usepackage{booktabs}
\usepackage{natbib}
\usepackage[colorlinks=true,linkcolor=blue,citecolor=blue,urlcolor=blue]{hyperref}
\usepackage{microtype}
\title{The Drift Contract: Spectral Updates for Depth-Robust Local Learning}
\author{Fabien Polly\\ \small Independent researcher}
\date{July 2026}

\newcommand{\eps}{\epsilon}

\begin{document}
\maketitle

\begin{abstract}
Local learning trains each layer with its own auxiliary loss and no global backward pass, which makes layer updates structurally parallel. Two problems have kept it marginal: accuracy degrades as depth grows, and hyperparameters are fragile. We apply Muon-style spectral update geometry (momentum orthogonalization with spectral step scaling) to per-layer local updates, an intersection not previously studied. On CIFAR-10 MLP benchmarks with local linear heads, a single step-size setting is the best value in our tested grids from width 128 to 2048 and from depth 12 to 48, while local Adam requires re-tuning along both axes and still collapses at depth 48 (31.3 percent re-tuned per depth, 19 percent with its depth-12 setting transferred, vs 42.7 percent for the spectral update at its unchanged setting). At five seeds and width 512 the spectral update leads local Adam by a clear margin ($48.9 \pm 0.5$ vs $46.6 \pm 0.3$). Prospectively specified controls attribute the transfer and most of the depth robustness to the spectral geometry itself rather than to any step-size rule on top of it. We additionally formulate the step size as a drift contract, $\mathrm{lr} = \eps / \mathrm{RMS}(\text{input})$, which bounds each layer's weight-induced pre-activation change per step, conditioned on its current input. The contract yields a small gain over the best fixed learning rate where that baseline is measured, makes the step size interpretable, and provides a per-layer, input-conditioned drift bound that standard optimizers do not offer. We report one negative result: with RMSNorm and weight decay in the trunk, the stability benefit of spectral updates accrues to global rather than local training, so the local advantage concentrates precisely where normalization is absent.
\end{abstract}

\section{Introduction}

Backpropagation couples every layer to the loss through a full backward pass. Local learning removes this coupling: each layer receives a gradient from a small auxiliary head and updates immediately \citep{nokland2019local,belilovsky2019decoupled,hinton2022forward}. The benefit is structural parallelism (4.34x throughput at 12 layers in our pipeline implementation, separate from this paper's code release); the historical costs are a depth pathology, where features degrade after the first few layers, and brittle hyperparameters.

Independently, Muon \citep{jordan2024muon} showed that constraining the spectral norm of weight updates (momentum, Newton-Schulz orthogonalization, spectral scaling) roughly halves the training cost of large transformers \citep{liu2025muon}. The theory behind it, steepest descent under operator norms, modular norms, and muP-style scaling \citep{bernstein2024old,large2024modular,yang2022tensor}, assumes gradients from a global backward pass. Whether any of it applies to local gradients was, to our knowledge, unexamined.

Why this matters beyond benchmarks: local training that survives depth, needs no per-layer tuning, and carries a per-step drift bound is well suited to settings where a global backward pass is impossible or unwanted, such as memory-constrained edge devices, decentralized pipelines over slow links, backprop-hostile hardware, and continually learning systems that should not change behavior abruptly.

This paper studies that intersection. Contributions:

\begin{enumerate}
\item First systematic study of spectral update geometry in local learning, with prospectively specified protocol and controls (fixed-lr attribution, auxiliary-parameter confound, normalization).
\item A depth result: local Adam degrades from 44.4 to 31.3 percent as depth goes 12 to 48 even with per-depth re-tuning; spectral updates lose 3.0 percentage points with a single unchanged setting.
\item A transfer result: one setting (3e-3) is the best-performing value in the tested grids across a 16x width range and a 4x depth range. The optimum is interior (accuracy collapses when the setting is 10x smaller and decays when larger) and lands at the same value on both axes.
\item The drift contract: expressing the step as $\mathrm{lr} = \eps / \mathrm{RMS}(\text{input})$ turns the learning rate into a per-layer, input-conditioned bound on weight-induced functional change (Section 2 states the bound and its slack exactly; orthogonalized updates have approximately unit spectral norm). It edges the best fixed lr where both are measured (about +1 point at width 1024), and it beats local Adam at every width: +1.2, +2.3, +2.9 points at widths 128, 512, 1024 (five seeds), and +1.0 at width 2048 (two seeds).
\item Negative and attribution results: RMSNorm plus weight decay moves the stability advantage to global Muon; width transfer is due to spectral scaling, not the contract; bias updates are not the source of the residual drift floor.
\end{enumerate}

\section{Method}

\paragraph{Setup.} A depth-$L$ MLP, ReLU, no trunk normalization unless stated. Layer $l$ has weights $W_l$, bias $b_l$, and a linear local head $H_l$ mapping its activation to class logits. Each layer minimizes its own cross-entropy $L_l$; gradients do not flow between layers. All layers update simultaneously from the same forward pass, matching a parallel pipeline execution.

\paragraph{Drift-bounded spectral update} (per layer, matrices only), with $\mu = 0.95$:
\begin{align}
m &\leftarrow \mu\, m + G \\
O &= \mathrm{NewtonSchulz5}(m) \\
W &\leftarrow W - \mathrm{lr}\, \sqrt{\max(1, d_{\text{out}}/d_{\text{in}})}\; O
\end{align}
$\mathrm{NewtonSchulz5}$ is five iterations of the standard quintic polynomial with coefficients $(3.4445, -4.7750, 2.0315)$ applied to $m$ normalized by its Frobenius norm; it approximately maps all singular values of $m$ to one using matrix products only \citep{jordan2024muon}. Biases use Adam with the update RMS clipped to $\eps$. Heads use Adam. This hygiene follows Muon practice: non-matrix parameters stay on elementwise optimizers.

\paragraph{Drift contract.} Newton-Schulz yields approximately unit spectral norm; write $\kappa_l = \lVert O \rVert_2$, close to 1 in practice. With the scaling above, $\lVert \Delta W \rVert_2 = \mathrm{lr}\, \kappa_l \sqrt{\max(1, d_{\text{out}}/d_{\text{in}})}$, and for input $x$:
\begin{equation}
\mathrm{RMS}(\Delta z) \;\le\; \mathrm{lr}\; \kappa_l\, \sqrt{\max(1, d_{\text{in}}/d_{\text{out}})}\; \mathrm{RMS}(x).
\end{equation}
For square and expanding layers ($d_{\text{out}} \ge d_{\text{in}}$, all layers here except the CIFAR input layer) the shape factor is 1, and setting
\begin{equation}
\mathrm{lr}_l = \frac{\eps}{\mathrm{RMS}_{\mathrm{ema}}(h_{l-1})}
\end{equation}
bounds the weight-induced pre-activation drift per step by $\eps\,\kappa_l$, conditioned on the current input and on the EMA tracking the batch RMS (a strict version would use $\max(\mathrm{RMS}_{\text{current}}, \mathrm{RMS}_{\mathrm{ema}})$). Contracting layers loosen the bound by $\sqrt{d_{\text{in}}/d_{\text{out}}}$ (about 4.9 for the CIFAR input layer); replacing the step scale by $\sqrt{d_{\text{out}}/d_{\text{in}}}$ without the max removes this slack for every shape. The bias update is budgeted separately at $\eps$, so total per-layer parameter-induced drift is bounded by about $2\eps$. This is a conditional, per-layer bound on parameter-induced drift; it does not cover drift propagated from simultaneous upstream updates, which our measurements show grows with depth. We verify the per-step bound directly (CIFAR, widths 128 and 512): the measured ratio $\mathrm{RMS}(\Delta z)/\eps$ has mean near one, the default max-scaled step leaves a heavy tail on the input layer (p99 up to 4.3, its $\sqrt{d_{\text{in}}/d_{\text{out}}}$ slack), and the exact-scaled variant holds the ratio to a p99 of 1.09 and a maximum of 1.19 across layers, so the strict form of the bound is essentially respected. $\eps$ is the only knob. In local learning the only inter-layer obligation is the stability of a layer's output for its immediate consumer, so this per-pair bound replaces the global norm composition that modular-norm methods require \citep{large2024modular}.

\section{Experimental setup}

CIFAR-10, 20000 train / 5000 test, images flattened to 3072 features, normalized with train statistics. MLP width 128 to 2048, depth 12 to 48, batch 256, 12 epochs, He init. Auxiliary parameters under Adam with lr 3e-3 in every arm (an earlier protocol tied them to the swept lr in Adam arms only; we identified and removed this confound). Divergence detection covers activations, not only weights. Metrics: test accuracy; activation drift, measured as the relative RMS change of every layer's activations on a frozen probe batch over windows of training steps; and effective rank of features, defined as the exponential of the entropy of the normalized squared singular values of the activation matrix. Seeds: 2 for sweeps, 5 for headline configurations. Synthetic-task stability results (Section~\ref{sec:stability}) use a 10-class teacher-warped Gaussian mixture. Protocol, criteria, and both possible outcomes of each control were written down before running (prospectively specified, with the record in the repository history).

Compute: all synthetic experiments run in about one hour on a laptop CPU (pure NumPy). CIFAR campaigns ran on one RTX 4070 Ti via the CuPy backend. CPU and GPU paths agree (0.949 vs 0.948 on the regression check).

\section{Results}

\subsection{Width transfer with an interior optimum}

\begin{figure}[t]
\centering
\includegraphics[width=0.62\linewidth]{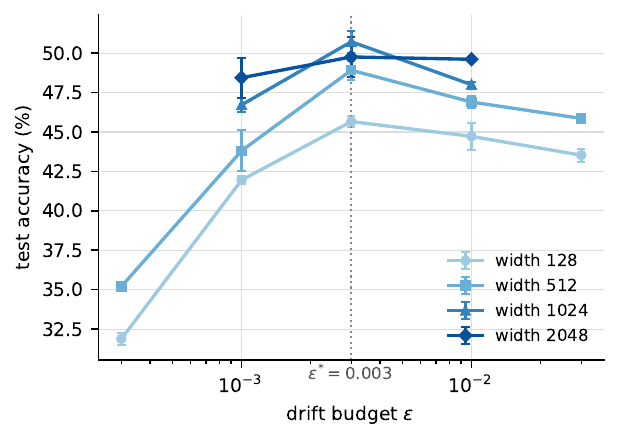}
\caption{Test accuracy as a function of the drift budget $\eps$ at four widths (error bars: one standard deviation over seeds). The optimum is interior and identical at every width, while Adam's optimal learning rate shifts from 1e-3 at width 128 to 3e-4 at width 512 and beyond.}
\label{fig:eps}
\end{figure}

Figure~\ref{fig:eps} shows contract accuracy by $\eps$. The curve is a full bell on every width with the peak at the same $\eps^{*} = 0.003$: accuracy collapses at $\eps = 0.0003$ (underfitting at fixed epochs), peaks at 0.003, and decays above. Adam's optimum moves: 1e-3 at width 128, 3e-4 at width 512, 1e-4 to 3e-4 at width 1024, 3e-4 at width 2048 with 2.6 points of seed spread. Measured drift at fixed $\eps$ is width-invariant within 15 percent.

\subsection{Depth: the main result}

\begin{figure}[t]
\centering
\includegraphics[width=0.62\linewidth]{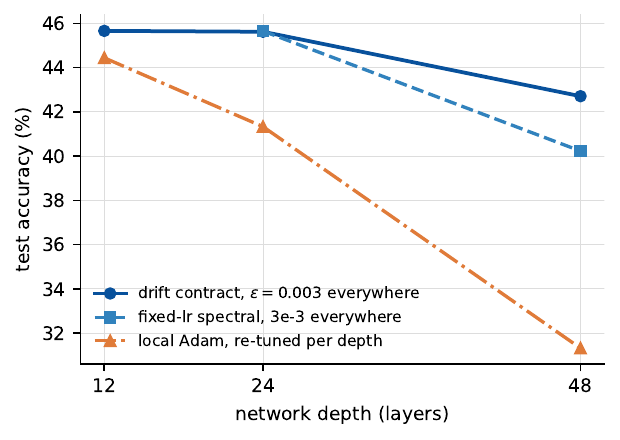}
\caption{Depth scaling at width 128, best setting per arm at each depth. Spectral arms keep one setting at all depths; Adam is re-tuned per depth and collapses anyway.}
\label{fig:depth}
\end{figure}

\begin{table}[t]
\centering
\caption{Depth scaling at width 128, test accuracy. The spectral arms keep
one transferred setting ($\eps=0.003$; fixed-lr $3e{-}3$); Adam is re-tuned per
depth. Depth 12 is 5 seeds; depths 24 and 48 are 2 seeds for the spectral arms.}
\label{tab:depth}
\begin{tabular}{lccc}
\toprule
Depth & Contract ($\eps$ 0.003) & The fixed-lr spectral update (3e-3) & Local Adam (best lr) \\
\midrule
12 & 45.7 & -- & 44.4 \\
24 & 45.6 & 45.6 & 41.3 \\
48 & \textbf{42.7} & 40.2 & 31.3 \\
\bottomrule
\end{tabular}
\end{table}

Table~\ref{tab:depth} and Figure~\ref{fig:depth}: Adam loses about 13 points over the depth range despite per-depth re-tuning (its width-128-optimal setting scores 19 percent at depth 48), while the contract loses about 3. Spectral arms keep one setting throughout. $\eps^{*} = 0.003$ at all three depths. The fixed-lr spectral entry at depth 12 is left blank because the width-128 base has no fixed-lr run in this campaign.

\subsection{Five-seed statistics}

\begin{table}[t]
\centering
\caption{Headline configurations, validation-selected, test accuracy, mean
$\pm$ standard deviation. Widths 128, 512, 1024 are 5 seeds; width 2048 is 2
seeds (outside the seed sweep).}
\label{tab:seeds}
\begin{tabular}{lccc}
\toprule
Width & Contract & Local Adam & Gap \\
\midrule
128 & $45.66 \pm 0.32$ & $44.44 \pm 1.49$ & +1.22 \\
512 & $48.92 \pm 0.55$ & $46.59 \pm 0.31$ & +2.33 \\
1024 & $50.72 \pm 0.63$ & $47.84 \pm 0.71$ & +2.88 \\
2048 & $49.76 \pm 0.90$ & $48.78 \pm 1.86$ & +0.98 \\
\bottomrule
\end{tabular}
\end{table}

Table~\ref{tab:seeds}: the contract beats local Adam at every width, by 1.2 to 2.9 points on the five-seed widths (128, 512, 1024). The width-2048 gap (+1.0) rests on two seeds and Adam's larger seed variance there, so we do not read a trend into it. The clean separation is at widths 512 and 1024, where the five-seed means are well apart.

\subsection{Stability and the normalization caveat}
\label{sec:stability}

On the synthetic task without trunk normalization, the local spectral arm is stable over a 333x learning-rate range (no divergence up to lr 1.0) vs 100x for local Adam and 10x for both global arms; the local-times-spectral interaction is super-additive. Adding RMSNorm and decoupled weight decay reverses the picture on the spectral side: global Muon becomes the most stable arm (300x) while the local spectral arm narrows to 30x. Local arms still degrade gracefully where global Adam fails abruptly. We conclude the stability benefit of spectral local updates is largest exactly where normalization is absent, and report this as a scope condition rather than hiding it.

\subsection{Attribution controls}

The fixed-lr spectral update transfers its optimum (3e-3) across all four widths and both extra depths, so setting transfer belongs to the spectral scaling, consistent with muP-style analyses of Muon in global training \citep{bernstein2024old,yang2022tensor}. At depth 24 the fixed-lr spectral update matches the contract (45.6 vs 45.6); at depth 48 the contract leads by 2.5 points (42.7 vs 40.2, two seeds), and the fixed-lr window narrows with depth (lr 1e-2 scores 44.0 at depth 12 but 29.1 at depth 48). The contract's specific added value is: a small gain over the best fixed lr where measured (about 1 point at width 1024), the per-layer drift bound, and per-layer adaptation to input scale. Bias ablations (frozen, normalized momentum, clipped Adam) show biases do not cause the residual drift floor; constant-norm orthogonalized steps do, since they never anneal.

\subsection{Mechanism}

Effective rank of layer-12 features at matched settings: 27 to 47 under spectral updates vs 6 to 23 under Adam (width 512 and up). Per-layer probes show local Adam peaks at layer 2 and degrades monotonically after, while spectral arms stay flat through depth. This matches the gradient-spectrum picture reported for global ViT training \citep{southworth2026muon}: Adam concentrates gradient energy in a narrow basis, orthogonalization spreads it. Whether the flattened spectrum itself is the causal ingredient is debated: random or inverted spectra can match Muon in global training, with step alignment and magnitude proposed as the operative factors \citep{shumaylov2026muon}. Our claims therefore rest on the measured transfer and depth results, not on a mechanistic commitment; spectrum ablations (random spectrum, polar factor, normalization-only) are future work.

\section{Related work}

Muon and its analysis: orthogonalized momentum and spectral-norm-constrained steepest descent \citep{jordan2024muon,bernstein2024old}, scaled to LLM training \citep{liu2025muon}, studied spectrally in vision transformers \citep{southworth2026muon}, generalized as layer-wise LMO optimization \citep{riabinin2025gluon}, and challenged mechanistically \citep{shumaylov2026muon}. Learning-rate transfer: muP \citep{yang2022tensor} and the modular norm \citep{large2024modular} obtain width-transferable learning rates in global training via norm composition over the full graph. Closest to our transfer result, \citet{ishikawa2024local} derive muP parameterizations for local losses (predictive coding and target propagation) and demonstrate width transfer of learning rates; our setting differs in the local rule (auxiliary classification heads), in extending the transfer to depth, and in deriving the step size from a per-layer drift bound rather than from a parameterization. Local and greedy layer-wise learning: auxiliary local losses \citep{nokland2019local}, decoupled greedy learning \citep{belilovsky2019decoupled}, and forward-forward \citep{hinton2022forward} established the paradigm and its depth plateau. To our knowledge no prior work applies spectral update geometry to auxiliary-head local gradients or derives step sizes from an inter-layer drift bound.

\section{Limitations}

MLP classifiers on CIFAR-scale data; linear local heads; 12 fixed epochs for the headline. A 50-epoch check at width 128 shows local Adam roughly closing the width-128 gap (both near 44 to 46 percent), so that gap is largely an optimization-speed effect rather than a converged-accuracy one; the depth collapse, by contrast, persists. Per-layer drift is bounded but total network drift grows with depth (0.20 to 0.55 from depth 12 to 48 at fixed $\eps$, suggesting depth-aware budgets); Newton-Schulz overhead is 10 to 15 percent at these sizes. Hyperparameters are selected on a held-out validation split and test accuracy is reported at the selected setting; width 2048 and two Adam configurations use two seeds rather than five (outside the seed sweep). The drift bound is conditional (Section 2): it covers weight-induced drift at the current input, not propagated or off-distribution drift. The negative normalization result bounds the claim: spectral local updates matter most in unnormalized or normalization-averse settings (analog hardware, minimal-memory edge training). $\eps$ is scale-invariant but not domain-invariant (0.01 on our synthetic task, 0.003 on CIFAR).

\section{Conclusion}

Spectral update geometry removes the two historical costs of local learning on our benchmarks: depth degradation and hyperparameter fragility. One number, 3e-3, read either as a fixed spectral step or as a drift budget, works from width 128 to 2048 and depth 12 to 48. The drift-contract reading adds a property no standard optimizer provides: a per-layer, input-conditioned bound on how far each step can move the layer's output.

For systems that retrain without a human in the loop, the two properties compose usefully. Such a system no longer needs its learning speed hand-tuned: the speed follows from a change budget fixed once per domain, and that budget caps each layer's own per-step contribution to behavioral change (a conditional bound; drift propagated from upstream layers is not covered). Tuning optimizes the loss; the contract bounds the risk, and unattended retraining needs both. The exact promise, on our benchmarks, is: calibrate $\eps$ once per domain on a small model, then reuse it unchanged as the model grows or deepens.

For decentralized training, our results lift the two locks that belonged to optimization, depth degradation and per-node tuning, and leave standing the two that do not: forward-pass network latency, and the quality of local losses at language-model scale. We consider the latter the natural next question. Local learning with contracts starts to look like what it structurally is: a set of autonomous workers bound by simple bilateral agreements.

\section*{Reproducibility}

All code is pure NumPy (optional CuPy backend), a single library file plus one runner per experiment, JSON checkpoints after every run. Synthetic results reproduce in about one hour on a laptop CPU. Code and raw results: \url{https://github.com/infinition/drift-contract}.

\end{document}